\documentclass{article}

\usepackage{arxiv}

\usepackage[utf8]{inputenc} 
\usepackage[T1]{fontenc}    
\usepackage{hyperref}       
\usepackage{url}            
\usepackage{booktabs}       
\usepackage{amsfonts}       
\usepackage{nicefrac}       
\usepackage{microtype}      
\usepackage{lipsum}		
\usepackage{graphicx}
\usepackage{natbib}
\usepackage{doi}
\usepackage{multirow}

\title{Meta-Learning Approaches for a One-Shot Collective-Decision Aggregation: \\
Correctly Choosing how to Choose Correctly}

\date{} 					

\author{ 
Hilla Shinitzky, Yuval Shahar, Ortal Parpara, Michal Ezrets and Raz Klein \\
The Department of Software and Information Systems Engineering  \\
Ben-Gurion University of the Negev \\
Beer-Sheva, Israel\\
Corresponding author: \texttt{hillash@post.bgu.ac.il}\\
}

\renewcommand{\shorttitle}{Meta-Learning Approaches for a One-Shot Collective-Decision Aggregation}

\hypersetup{
pdftitle={Meta-Learning Approaches for a One-Shot Collective-Decision Aggregation},
pdfauthor={Hilla Shinitzky},
}

\begin{document}
\maketitle

\begin{abstract}

Aggregating successfully the choices regarding a given decision problem made by the multiple collective members into a single solution is essential for exploiting the collective's intelligence and for effective crowdsourcing. There are various aggregation techniques, some of which come down to a simple and sometimes effective deterministic aggregation rule. However, it has been shown that the efficiency of those techniques is unstable under varying conditions and within different domains. Other methods mainly rely on learning from the decision-makers previous responses or the availability of additional information about them. In this study, we present two one-shot machine-learning-based aggregation approaches. The first predicts, given multiple features about the collective's choices, including meta-cognitive ones, which aggregation method will be best for a given case. The second directly predicts which decision is optimal, given, among other things, the selection made by each method. We offer a meta-cognitive feature-engineering approach for characterizing a collective decision-making case in a context-sensitive fashion. In addition, we offer a new aggregation method, the \textit{Devil's-Advocate} aggregator, to deal with cases in which standard aggregation methods are predicted to fail. 
Experimental results show that using either of our proposed approaches increases the percentage of successfully aggregated cases (i.e., cases in which the correct answer is returned) significantly, compared to the uniform application of each rule-based aggregation method. We also demonstrate the importance of the Devil's Advocate aggregator. 
\end{abstract}


\section{Introduction}

At this time, there is a full awareness of the wisdom of the crowd effect \citep{surowiecki2005wisdom} and the collective's intelligence. A nominal group of individuals can often achieve a better outcome in a broad spectrum of tasks than one individual alone \citep{surowiecki2005wisdom}. With the help of technological solutions and the current global interconnected world, it has become easier and popular to utilize crowds' wisdom, such as in crowdsourcing applications \citep{howe2006rise}. 

Examples of crowdsourcing tasks and applications include solving temporal ordering problems in a distributed fashion \citep{steyvers2009wisdom}, political and economic forecasting \citep{budescu2014identifying, mellers2014psychological}, and evaluating information retrieval systems \citep{alonso2008crowdsourcing,lease2012crowdsourcing}.

In this context, an important issue to be dealt with is how the final result is determined. Composing methods for combining the perspectives of multiple individuals into one collective decision is an essential issue in various research communities. 

Since there is a growing necessity for reliable aggregation methods in crowdsourcing systems, most effort is invested in learning the workers' quality, using their past performance in different tasks. Usually, the assessed quality of the workers is used to weigh their contribution to the crowd's final aggregated decision. In general, most of the more sophisticated aggregation techniques rely on workers' track records or the availability of additional information related to the individuals (multiple responses, demographic information, etc.) \citep{welinder2011multidimensional,bachrach2012crowd,bachrach2012grade,gaunt2016training,laan2017rescuing,zheng2017truth,weld2015artificial}. 
The effectiveness of approaches of these types is compromised in challenging or unique cases where the individuals' quality is not enough to ensure a successful aggregation, and in situations where the necessary information is unavailable. 

Another group of fairly popular and well-known methods does not use any information beyond the responses to the relevant task, and comes down to a simple and sometimes effective deterministic aggregation rule \citep{koriat2008subjective,koriat2012two,hertwig2012tapping,prelec2017solution,hastie2005robust,kerr2004group,sorkin1998group,surowiecki2005wisdom}. 
While simplicity can be a significant advantage, it has been shown that different techniques are preferable in different situations \citep{bahrami2010optimally,koriat2016views,lee2012inferring,chen2004eliminating,koriat2018prototypical,lorenz2011social,simmons2010intuitive,levine2015social}. 
One may ask themselves, would it not be beneficial to find a way to know which method to use in each given case?

Our approach attempt to gain "the best from both worlds" - applying a sophisticated and flexible aggregation technique without dependency on information about the decision-makers or the task. This is being done through: 1) learning from and on the simple rule-based methods' successes and failures, 2) learning from the context provided by the set of responses, i.e., consider the behavior of the group, rather than the specific individual's. 

We propose two one-shot (i.e., without relying on any knowledge regarding the involved specific group's or specific individuals' performance in the past) machine-learning-based aggregation approaches that exploit the context created by the various properties of the problem (task) and the associated responses. At the core of each approach lies a classification model that learns from other past aggregation cases and from their associated features. 
Our first approach uses a classification model to predict which \textit{aggregation technique} will prove the best to apply, and our second approach focuses on predicting directly which \textit{answer} is the correct answer.

Another contribution of our work is a new aggregation method created to function as a \textit{Devil's Advocate}, which selects the option that is as far as possible from the other methods' choices. The importance of this method is \textit{not} in its performance as a stand-alone aggregation method; it is unlikely to achieve a high success rate in a large variety of tasks. The new aggregation method's strength relies on its ability to enable the reconsideration of alternatives that otherwise will be overlooked, which, in challenging decision-making tasks, could turn out to be the correct decisions. 

Our approaches are unique, first of all in their purpose, to observe and learn from group-level features and to infer the best choice even in cases in which there is a high probability of a majority vote failure. 
Both of our approaches exploit unique meta-cognitive features, such as level of confidence in the answer and assessment of the popularity of the choice by other group members, and new features generated from the data through various feature-engineering techniques.

Moreover, in contrast to other existing machine-learning-based aggregation methods \citep{welinder2011multidimensional,bachrach2012crowd,bachrach2012grade,gaunt2016training,laan2017rescuing,zheng2017truth,weld2015artificial}, which often try to learn over time “who are the experts in the group”, our approaches have the advantage of being “one-shot” techniques. Our methodology is thus independent of the crowd-specific composition, prior knowledge and personal track records. It is adaptive to various situations (contexts) in an automated fashion. 
Thus, the concepts underlying our approaches are not confined to specific tasks, and can be generalized to any purpose in which the end goal is to perform a successful aggregation process.

\section{Related Work}

Combining decisions made by multiple individuals in a group setting or as a dispersed crowd (i.e., crowdsourcing) has been shown to improve the quality of the final decision, relative to the average performance of an individual, in various domains, tasks, and environments. This phenomenon has also been referred to as the "wisdom of the crowd" \citep{surowiecki2005wisdom}. 
While some studies attempted to find the conditions under which the wisdom of the crowds holds \citep{davis2014crowd,davis2015composition}, other studies addressed the challenge of improving decisions quality made by a group of individuals while inspecting different aspects in the process. 
A major line of research attempts to improve the quality of the final decision by applying different methods for aggregating multiple decisions into one collective decision. 
The most common aggregation method is the majority rule (MR), which selects the decision that the majority of individuals agree on. This method can be implemented easily and can accurately aggregate a collection of decisions in a non-negligible portion of the time \citep{hastie2005robust,kerr2004group,sorkin1998group}.  

That been said, there is evidence for the failure of the majority rule and democratic voting in general, in different forms of interactions and group settings \citep{chen2004eliminating,koriat2018prototypical,lorenz2011social,simmons2010intuitive}.
Moreover, there are still several types of decisions in which the group performance does not surpass individuals' performance. In some cases, especially when some form of interaction between group members, or at least an inspection of other group members' solutions, is allowed, groups can even become \textit{less effective} as the size of the group \textit{grows} \citep{amir2013verification,amir2018more,galesic2018smaller}. This paradoxical deteriorating performance phenomenon could be due to a variety of reasons, such as the computational characteristics of the given problem \citep{amir2013verification}, low demonstrability of the solution of strategic problems to other group members\citep{amir2018more}, negative effects of group interactions, and more \citep{bahrami2010optimally,koriat2018prototypical,laughlin1986demonstrability,laughlin2006groups,levine2015social}. 

With that in mind, collective intelligence can be exploited optimally only by understanding its limitations and the defining conditions. The task's characteristics and domain, the group composition, and the aggregation technique are all defining factors for the collective's success. 
In some tasks, such as particularly challenging tasks or those requiring expertise, it could be preferable to be more selective with the individuals on whom the final decision relies.
Some studies examined the possibility of identifying experts or more trustworthy individuals and aggregate the final decision by taking their perspective only \citep{goldstein2014wisdom,mannes2014wisdom}.

Another approach is to use a weighted-aggregation rule, which considers each vote but with a varying impact on the final result (i.e., a weight). A typical approach is to base the weight of a respondent's vote on their abilities. That is, to identify the solid individuals or experts and, instead of relying solely on them, to magnify their impact on the outcome \citep{budescu2014identifying,lee2012inferring,yue2014weighted}. For example, \citet{yue2014weighted} proposed a weighted aggregation rule for tasks that has low-agreement among the crowd. The weight of each answer is based on the respondent's past performance. 
Note that such methods require additional information about the individuals, such as historical records on past performance, to evaluate the individuals' abilities and expertise. We wish to avoid using this kind of information in our work, so there will be no dependencies on extra data aside from the collective's responses to the given problem.

There are other types of information that can be used for a weighted-aggregation technique, such as the subjective-confidence, i.e., the reported confidence of the individual about their decision \citep{koriat2008subjective,hertwig2012tapping}. Subjective confidence can be used not only through weighting but also through maximization, i.e., to select the decision with the highest average confidence reported by its supporters \citep{koriat2008subjective,koriat2012two,hertwig2012tapping}. 
While performing well in some cases, the confidence-based methods are not always dependable and suited for specific domains \citep{bahrami2010optimally,hertwig2012tapping,koriat2016views,lee2012inferring,aydin2014crowdsourcing,koriat2008subjective,koriat2012two}. 

The idea to use reflective, meta-cognitive information provided by the individual on themselves, such as the confidence-based methods, is the Surprisingly-Popular (SP) method \citep{prelec2017solution}. The SP method consists of choosing the answer that has had surprisingly large support, relative to the mean predicted support to that answer as estimated by the participants, and not necessarily the majority’s opinion. The method fared well in the researchers’ experiments, though mostly in common-knowledge binary questions. The SP method is designed to be especially useful for cases in which the problem is misleading and tends to elicit mostly wrong responses, and the confidence-based and majority-rule usually fail.  
Several studies examined the performance of SP under different conditions and different domains of tasks \citep{rutchick2020does,lee2018testing,gorzen2019extracting}.
Across various studies, the comparison of aggregation performance by SP, confidence-based methods, and majority rule showed no consistent conclusion or definitive superiority of one method.  
Inspired by the idea behind SP, other works proposed other types of aggregation techniques that involve a similar prediction mechanism to apply on specific tasks. For example, \citet{martinie2020using} proposed the Meta-Probability Weight (MPW) algorithm for answer-aggregation in multiple-choice forecasting questions.

While offering an easy to apply one-shot aggregation procedure, the simple rule-based methods are also occasionally unreliable. So, the use of sophisticated model-based and statistical methods, as well as Machine-Learning techniques \citep{welinder2011multidimensional,bachrach2012crowd,bachrach2012grade,gaunt2016training,laan2017rescuing,zheng2017truth,weld2015artificial}, has been attempted in the context of crowdsourcing and aggregating a collective choice. 
Similar to experts identification methods, the sophisticated, model-based, and Machine-Learning based methods rely on past performance or an array of responses associated with each individual to evaluate their contribution to the collective performance or to learn their capabilities and expertise. 
Thus, even though these methods can achieve better performance than majority-vote, they require additional information and do not offer a one-shot aggregation such as our approach. 

\section{Definitions}

We focus on the general form of collective decision-making procedure. There is a problem (i.e., task) to be solved, several different options (answers) to choose from, and only one is considered the optimal or correct solution. 
Each member provides their choice regarding the solution to the given problem in the form of a response. The goal is to arrive at the optimal solution based on the perspectives of the entire group, i.e., a set of responses. 

Formally, we define a \textit{collective decision-making case} by the set of responses $R_P$, associated with the decision-making problem $P$, its set of possible answers $A_P$, and one optimal answer denoted by $a^*$. 
Table \ref{tab:response-table} details the format of a response in our setting.  
Next, we define some values aggregated from $R_P$.

We refer to the sub-set $R_{P(a_i)}\subseteq R_P$ as the \emph{supporters} of answer $a_i$, i.e., the responses of those who voted for $a_i$
The support rate of answer $a_i$ (i.e., popularity), denoted by $S(a_i)$, is the percentage of votes for $a_i$, and $S = \{ S(a_i)|\forall{a_i\in A_P}\}$ denotes the support rate distribution.
Let $a_{max}$ denote the most popular answer (i.e., the answer that received the highest percentage of votes), and $a_{min}$ denote the least popular answer. Thus, $S(a_max)$ and $S(a_min)$ are the support rate of the most popular, and least popular answer, respectively. 

The set of confidence reports denoted by $C = \{r_j.c|\forall{r_j\in R_P}\}$. The set of confidence reports considering only the supporters of $a_i$ is given by $C(a_i) = \{r_j.c|\forall{r_j\in R_{P(a_i)}}\}$.
The set of predictions about the support of $a_i$, given by $PS(a_i) = \{r_j.ps[a_i]|\forall{r_j\in R_P}\}$.

Now, on a given set of values, such as the ones defined above, we can apply variety of functions: $Average(\cdot)$ for calculating a set's average; $Var(\cdot)$ for calculating a set's variance; $Max(\cdot)$ for extracting the maximal value; $Min(\cdot)$ for extracting the minimal value; and $Entropy(\cdot)$ for calculating the entropy. 

The process of selecting a final group decision, based on the group members' perspectives, is called an aggregation process, and can be performed using different aggregation methods.
We define an aggregation method as a function, which, given a set of responses for problem $P$, returns an answer $a_i$ from the set $A_P$. 
Assuming an aggregation method called $AG$ and the set $R_P$, the answer returned by $AG$ when applied over $R_P$ is given by $f_{AG}(R_P)$. The outcome of the aggregation procedure, denoted by $O_{AG}\in \{0,1\}$ represents its success (1) or failure (0), such that: $O_{AG}(R_P) = 1$ if $f_{AG}(R_P) == a^*$. 

The following are some relevant examples of aggregation method:
\begin{itemize}
\item \textbf{Majority Rule (MR):} Returns the answer that received the highest support ($a_i$ that maximize the value $S(a_i)$). 
\item \textbf{Weighted Confidence (WC):}  Returns the answer with the highest value of confidence-weighted support ($a_i$ that maximize the value $S(a_i)\cdot Average(C(a_i))$).
\item \textbf{Highest Average Confidence (HAC):} Returns the answer that received the highest support ($a_i$ that maximize the value $Average(C(a_i))$).
\item \textbf{Surprisingly Popular (SP):} Returns the answer that was more popular than predicted on average by respondents ($a_i$ that maximize the value $S(a_i)-Average(PS(a_i))$).
\end{itemize} 

\begin{table}[t]
 \caption{Response format}
  \centering
  \begin{tabular}{ll}
    \toprule
    Value     & Description     \\
    \midrule
    $v$ & A vote for an answer in $A_P$       \\
    $\left \{ ps[a_1],...,ps[a_m] \right \}$     & Predicted support for each $a_i \in A_P$      \\
    $c$     & Reported confidence         \\
    \bottomrule
  \end{tabular}
  \label{tab:response-table}
\end{table}

\subsection{The Devil's Advocate Aggregator}

One of the novelties we offer in this paper is a new aggregation method that functions as a Devil's Advocate to the rest of the aggregation methods. We, therefore, refer to this aggregation method as the \textit{Devil's Advocate} (DA) aggregator, and its purpose is to offer the opposition's perspective relative to a set of aggregation results. Specifically, the DA selects the answer chosen by the smallest number of aggregation methods in a process that somewhat resembles a "reverse majority rule" applied to an ensemble of classifiers.

Thus, contrary to other aggregation methods, the DA does not base its decision on the properties of the set of responses $R_P$, at least not directly. Its choice is based on the selections made by \textit{other} aggregation methods, while considering every option available in $A_P$ (even those that did not receive \textit{any} votes in $R_P$). 
Formally, given a set of answers options $A_P$ and a set of aggregation results ${f_{AG_1}(R_P),..,f_{AG_q}(R_P)}$, the DA aggregation procedure is performed as follows: 
For each $a_i$ in $A_P$ maintain a counter (initiated with 0), and increase the counter by 1 for each $f_{AG_j}(R_P)$ that holds $f_{AG_j}(R_P) == a_i$; Finally, return the answer with the lowest counter value.
There are several possible scenarios in which multiple answers are equally unpopular in the aggregation methods' choices. 
Thus, in some cases, depending on the size of $A_P$ and the number of aggregation methods considered, there is a need for a \textit{tie-breaker} methodology. In general, especially when $|A_P|>2$, we consider the DA tie-break method a specific implementation-design choice.

In our implementation, the DA aggregator's input includes the aggregation results of the MR, WC, HAC and SP methods, i.e., $\{f_{MR},f_{WC},f_{HAC},f_{SP}\}$. The tie-breaker methodology our implementation used  was selecting the answer that was not chosen by MR (and thus maintaining DA's role as an opposition); since we evaluated our approaches on binary-choice problems, this was sufficient. 

The importance of the DA does not rely on the accuracy of its aggregation performance when used indiscriminately (i.e., uniformly on all cases) but rather on its ability to allow the reconsideration of alternatives that otherwise will be overlooked, which, in challenging decision-making tasks, could turn out to be the correct decisions. 
In procedures of moderating between different aggregation methods, the DA can be crucial for their success. Obvious scenarios that demonstrate the importance of DA are cases in which no method manages to identify the correct answer.

\section{Answer-Aggregation through Learning and Classifying Decision-Making Cases}

We propose two approaches for aggregating the final decision of a collective based on the prediction of a trained Machine-Learning model. The first approach focuses on the selection of the optimal \textit{aggregation method} suited for the current group decision-making instance (e.g., the MR method); we refer to it as the \textit{Aggregation-Method-Prediction} (AMP) approach. The second approach focuses on the selection of the \textit{optimal answer}, based on primarily similar features (e.g., the second of two possible answers to the problem); we refer to it as the \textit{Direct-Answer-Prediction} (DAP) approach. 

Each instance in the data on which the models are trained, in both approaches, represents a \textit{collective decision-making case}. The difference between the two approaches and their associated Machine-Learning models, are the target variables (i.e., the models' output) - i.e., we are using essentially the same data for two different classification tasks. 

The overall procedure to be held when using both of our proposed approaches is composed of three steps:
The first step consists of training a classification model, using resolved cases of collective decision-making problems, represented by a set of sets of responses, $\{R_{P_1},...,R_{P_n}\}$, by generating a set of instances, $\{{I_1},...,{I_n}\}$ (one instance for each set of responses). 
The specific features extracted from each set of responses, which comprise the generated decision-making instance, are described in detail in the following subsection.
After training the classification model, it can be used for classifying new unseen cases, for the purpose of generating its aggregation result. 
Specifically, the trained model is used to obtain a prediction for a new collective decision-making case instance $I_{new}$, generated from a new set of responses $R_{P_{new}}$.
The final aggregation result is determined according to the appropriate classification prediction. In the case of choosing the optimal aggregation method, the predicted method will be applied to $R_{P_{new}}$, i.e., to the set of responses. In the case of determining the optimal answer, the predicted answer will be returned.

After laying down the general structure of our proposed approaches, and before we describe each in detail, we first address the composition of features that make up the collective decision-making case instance.
Note that in the case of our direct answer-prediction approach, the set of features, in addition to those used by the method-aggregation prediction approach, will also include the aggregation results, i.e., $f_{AG}$, the answer returned by applying each of the aggregation methods $AG$ that we are considering.

\subsection{Features-Engineering for a Collective Decision-Making Case Instance}
    
A fundamental step in developing our approaches is to compose features that characterize a collective decision-making case.  
Each set of responses $R_P$ associated with one problem is transformed into one instance $I$ (data row) to characterize a collective decision-making case. 
The task is to extract from $R_P$ as much information as possible that would reflect the context of the case, such as the complex nature of $P$, but without having any knowledge of the problem $P$, respondents' identity or associated information, nor their performance history. 

Since generating all of the features relevant to both approaches is not a trivial feature-engineering process, we used various strategies and techniques and incorporated knowledge derived from appropriate research fields.  

First, the more prominent values to incorporate are the aggregation criteria associated with standard aggregation methods. For instance, the Majority-Rule (MR) method selects the answer with the highest support (i.e., the most popular). Thus, some features could incorporate the value $S(a_{max})$ in one way or another.

Our second approach focuses on each component in each response separately (i.e., the chosen answer, the predictions, and the confidence level) and extracts the information given by its behavior and distribution across the entire set of responses. For example, a feature derived from the distribution of confidence levels can be the percentage of responses with high confidence.

In addition, each value extracted from the full set of responses of a given problem can be processed again, given a sub-sets of these responses. So, another approach to composing features is by observing the behavior and changes of those values across different sub-sets of responses to the same problem. 
The general process is executed as follows: 1) sample sub-sets of responses; 2) extract the desired information from each sub-set (e.g., the support rate of the most popular answer); 3) compose features given the set of values extracted from each sub-set (e.g., the variance of the support rate of the most popular answer, across sub-sets).
For simplification, we use the general notation $SG$, which denotes a set of specified values extracted from the sub-sets sampled from $R_P$. For example, $SG.S(a_{max})$ refers to the set of values of $S(a_{max})$ (the support rate of the most popular answer) across the sub-sets of responses. 

The set of engineered features is listed in Table \ref{tab:features}. 

\begin{table}[ht]
\caption{Collective Decision-Making Case's Engineered Features Table}
\centering
\begin{tabular}{lll}
\hline
  & \textbf{Feature Name} & \textbf{Description}  \\ \hline

1 & \textit{N}          & $|R_P|$   \\ \hline

2 & \textit{D\_Smax\_Smin}   & $S(a_{max})-S(a_{min})$ \\ \hline

3 & \textit{ES}          & $Entropy(S)$ \\ \hline

4 & \textit{VarS}        & $Variance(S)$ \\ \hline

5 & \textit{D\_S\_Uniform}   & Vector distance of $S$ from uniform distribution \\ \hline

6 & \textit{CAmin}    & $Average(C(a_{min}))$   \\ \hline

7 & \textit{CAmax}     & $Average(C(a_{max}))$   \\ \hline

8 & \textit{MinC}          & $Min(C)$ \\ \hline

9 & \textit{AvgC}       & $Average(C)$ \\ \hline

10 & \textit{VarC}   & $Variance(C)$ \\ \hline

11 & \textit{D\_MaxC\_AvgC}        & $Max(C)-Average(C)$   \\ \hline

12 & \textit{B\_Amax\_MaxCa}       &1, if $Average(C(a_{max})==Max(\{Average(C(a_i)|\forall{a_i\in A_P}\})$; 0, else \\ \hline
                                          
13 & \textit{MaxPSa}          & $Max(\{Average(PS(a_i))|\forall{a_i\in A_P}\})$ \\ \hline

14 & \textit{MinPSa}          & $Min(\{Average(PS(a_i))|\forall{a_i\in A_P}\})$  \\ \hline

15 & \textit{AvgPSv}          & $Average(\{r_j.ps[v]|\forall{r_j\in R_P}\})$  \\ \hline

16 & \textit{P\_lowC\_highPSv}         & Proportion of $r_j\in R_P$ such that $r_j.c<Average(C)$ and $r_j.ps[v]>AvgPSv$ \\ \hline

17 & \textit{P\_lowPSv\_highC}                & Proportion of $r_j\in R_P$ such that $r_j.c>Average(C)$ and $r_j.ps[v]<AvgPSv$  \\ \hline

18 & \textit{SG\_B\_Amax}     &1, if $a_{max}$ differs across sub-sets of responses; 0, else  \\ \hline

19 & \textit{SG\_VarSAmax}     & $Variance(SG.S(a_{max}))$ \\ \hline

20 & \textit{SG\_D\_MaxVarS\_MinVarS}         & $Max(SG.Variance(S))-Min(SG.Variance(S))$ \\ \hline

21 & \textit{SG\_D\_MaxES\_MinES}         & $Max(SG.Entropy(S))-Min(SG.Entropy(S))$ \\ \hline

22 & \textit{SG\_D\_MaxAvgC\_MinAvgC}    & $Max(SG.Average(C))-Min(SG.Average(C))$ \\ \hline

23 & \textit{SG\_D\_MaxVarC\_MinVarC}     & $Max(SG.Variance(C))-Min(SG.Variance(C))$  \\ \hline

24 & \textit{SG\_VarCAmin}       & $Variance(SG.Average(C(a_{min})))$ \\ \hline

25 & \textit{SG\_VarCAmax}     & $Variance(SG.Average(C(a_{max})))$  \\ \hline

26 & \textit{SG\_AvgCAmin}      & $Average(SG.Average(C(a_{min})))$ \\ \hline

27 & \textit{SG\_AvgCAmax}      & $Average(SG.Average(C(a_{max})))$  \\ \hline
\end{tabular}
  \label{tab:features}
\end{table}

Next, we provide specific descriptions of our proposed methods. 
As discussed earlier, the two methods use classification models, each trained on data sets of \textit{collective decision-making case} instances, where each instance includes the features described in Table \ref{tab:features}. The main difference between the two methods comes down to the classification task, i.e., different target variables. The final aggregation process is held based on the classifiers' predictions. 

While the technical differences are minor, the main ideas behind those two approaches are essentially different. One focuses on predicting which \textit{aggregation method}would be best if applied to a given case (and then applies it, to return the actual answer). The other takes a direct approach to predicting the correct \textit{answer}.  

\subsection{The Aggregation-Method-Prediction Approach}

Given a collective decision-making case, the \textit{Aggregation-Method-Prediction} (AMP) approach uses a classification model that predicts which \textit{aggregation method}, from a set of methods to consider $\{{AG_1},...,{AG_q}\}$ should be applied (i.e., which method has the highest probability of returning the correct decision). 
Given the model's prediction, the final aggregated answer is generated by applying the chosen aggregation method.
An input instance for the classification model is a \textit{collective decision-making case} instance, which includes the features described in Table \ref{tab:features}, and the classification target is the aggregation method to apply. 
Since there could be more than one method that can produce a successful aggregation outcome for a given case, the appropriate approach is to produce an independent prediction for each method in the form of a multi-label classification: 
Each method $AG_i$ has its correspondent binary-classification target label $O_{AG_i}$, which represents the outcome of applying it to the associated case. 
For example, consider the case $R_{P_j}$ and its associated instance $I_j$, and consider the labels $\{O_{MR}, O_{SP}\}$ which are assigned with the values $\{0,1\}$ (respectively). This means the SP method is predicted to produce a successful aggregation result when applied on $R_{P_j}$, while MR is predicted to fail. 

The full set of target labels we use in our implementation of AMP is $\{O_{MR}, O_{HAC}, O_{WC}, O_{SP}, O_{DA}\}$, representing the outcome of applying the aggregation methods MR, HAC, WC, SP, and DA, respectively. 

By definition, only one method is selected by the AMP approach to be applied to the group decision instance. However, due to our multi-label classification methodology, there might be more than one method whose predicted label's value is 1. 
To solve this dilemma, we use a binary classification that includes a \textit{confidence measure} (i.e., a classification probability), in which a predicted value of 1 results from a confidence measure $\in{(0.5,1]}$, and a predicted value of 0 results from a confidence measure $\in{[0, 0.5]}$.
Thus, given the predicted values for the aggregation methods $\{O_{AG_1},..., O_{AG_q}\}$ of a classified instance, the selected aggregation method ${AG_i}$ is the one whose label's value $O_{AG_i}$ has been assigned the highest confidence measure.

\subsection{The Direct-Answer-Prediction Approach}

The \textit{Direct-Answer-Prediction} (DAP) approach uses a classification model that predicts which \textit{answer} is the correct one and returns it as the final aggregated answer.
An input instance for the DAP approach is a \textit{collective decision-making case} instance, which includes not only all of the features described in Table \ref{tab:features}, but also the aggregation results (i.e., the chosen answer) when applying a set of aggregation methods - i.e., $\{f_{AG_i},...,f_{AG_q}\}$.
In our implementation, we included the aggregation results of the methods MR, HAC, WC, SP and DA - i.e., $\{f_{MR},f_{HAC},f_{WC},f_{SP},f_{DA}\}$.
The target classification label is the correct answer for the associated problem, i.e., $a^*$.
Thus, for binary-choice problems (i.e., when $|A_P| = 2$), the classification task here is a binary-classification task, and a multi-class classification task for multiple-choice problems (i.e., when $|A_P|>2$).
Note that due to the need for obtaining sufficient training data, we evaluated our proposed approaches on binary-choice problems only, but there is no technical limitation on applying our approaches to multiple-choice problems. 

\section{Evaluation}

\subsection{Material}

We used data collected from previous studies with open-access material \citep{martinie2020using,rutchick2020does,lee2018testing}.
These data sets contain responses to questions with two optional answers (i.e., binary decision problems), mainly associated with prediction problems (such as sports events outcomes and political elections) and general knowledge questions.

We excluded cases with either a 0\% or a 100\% correct solution rate.
The final data-set used in the evaluation procedure included 1209 decision-aggregation cases, with an average of 70 responses per case and an average of \%62 correct solution rate. 

The raw data set, composed of a set of responses $R_P$ for each problem $P$, was transformed into a data set in which each set $R_P$ was processed into one instance that included a set of features and target labels, according to the approach.

\subsection{Data set transformation}

The raw data set, composed of $n$ sets of responses $\{R_{P_1},...,R_{P_n}\}$, is transformed into a data set suited to the classification procedure performed in each approach (AMP and DAP), as described in the previous section. 
Each set $R_{P_j}$ is processed into one instance $I_j$, which include a set of features and target labels:
\begin{itemize}
    \item For the implementation of AMP, each instance $I_j$ include: (1) Features: collective-decision-making case instance features, listed in Table \ref{tab:features}. (2) Target labels: $\{O_{MR}, O_{HAC}, O_{WC}, O_{SP}, O_{DA}\}$ (the outcome of applying each aggregation methods: MR, HAC, WC, SP and DA).
    \item For the implementation of DAP each instance $I_j$ include: instance include: (1) Features: collective-decision-making case instance features, listed in Table \ref{tab:features} and $\{f_{MR},f_{HAC},f_{WC},f_{SP},f_{DA}\}$ (the selected answer by each aggregation method: MR, HAC, WC, SP and DA). (2) Target label: $a^*$ (the correct answer of $P_j$).
\end{itemize}

\subsection{Measures}

Since the main goal is to increase the likelihood of successful aggregation results, we measure each approach's performance by the rate of successful aggregation outcomes resulting from its respective classifier's predictions.

\subsection{Model Selection}

For each of our proposed approaches, we tested several classification techniques to be used at the final evaluation phase, according to the appropriate classification task:
\begin{itemize}
    \item The AMP approach includes a multi-label classification process \citep{zhang2013review}. There are three problem-transformation techniques used for implementing multi-label classification procedures: Binary Relevance (BR), Classifier Chain (CC), and Label Power-set (LP) \citep{boutell2004learning,read2009classifier,read2011classifier,read2008multi}.
    Each problem-transformation technique is a wrapper that transforms the classification problem into multiple multi-classification or binary-classification tasks, implemented by a given classification algorithm. Thus, the model selection procedure includes testing combinations of problem-transformation and classifiers together. We tested the three multi-label classification implementation techniques, BR, CC, and LP, over each of the following classifiers: Bernoulli Naïve Bayes (BNB), K-Nearest-Neighbors (KNN), Logistic Regression (LR), and Random-Forest (RF) (3X4= 12 classification techniques in total) \citep{mccallum1998comparison,manning2010introduction,altman1992introduction,bishop2006pattern,menard2002applied,breiman2001random}.
    \item The DAP approach includes a binary-classification process for binary-choice problems and a multi-class classification process for multiple-choice problems. For this approach, we tested the following classification algorithms: K-Nearest-Neighbors (KNN), Logistic Regression (LR), and Random-Forest (RF).
\end{itemize}

The performance of a classification model was measured by the rate of successful aggregation outcomes resulting from its predictions.

To select the best classification technique from the options tested for each of our proposed approaches to be used at the evaluation phase, we performed a nested cross-validation procedure as follows:
\begin{enumerate}
\item A 10-Fold cross-validation procedure: Split the data set into 10 sub-sets with equal size (each sub-set makes up 10\% of the data). 
\item For each sub-set $K\in [1,10]$, referred to as \textit{TestK}, perform the following procedure on the remaining 90\% of the data, referred to as \textit{TrainK}:
    \begin{enumerate}
    \item For each classification technique perform a leave-one-out cross-validation evaluation procedure:
        \begin{enumerate}
        \item For each instance $I$ in \textit{TrainK}:
            \begin{enumerate}
            \item Use all instances in \textit{TrainK} except $I$, to train a classification model according to the current classification technique. 
            \item Extract a prediction from the trained model, for instance $I$.
            \end{enumerate}
        \item Calculate the success rate achieved by applying the evaluated classification technique based on the predictions extracted in the previous step.
        \end{enumerate}
    \item Use all instances in \textit{TrainK} to train a classification model according to the classification technique that achieved the highest success rate in the previous step.
    \item Test the trained model by extracting predictions for every instance in \textit{TestK}.
    \item Calculate the success rate based on the predictions extracted in the previous step.
    \end{enumerate}
\end{enumerate}

For the AMP approach, the best overall performance was achieved by the Binary Relevance technique with the Random Forest classifier (for the full details of the model selection's results, refer to Table \ref{tab:cv-amp} in the Appendix). As for the DAP approach, the best results were achieved by the Random Forest classifier (for the full details of the model selection's results, refer to Table \ref{tab:cv-dap} in the Appendix). These classification techniques are used in the evaluation phase of our proposed approaches. 

\subsection{Final Evaluation Procedure - Prediction and Aggregation}

To evaluate each of our proposed approaches, we performed a leave-one-out procedure for extracting the model's predictions, using the classification techniques chosen in the model selection phase (BR+RF classifier for AMP, and RF classifier for DAP).
This process is done as follows: For each instance $I_j$ in the data set, train the classification model over the entire data set, excluding $I_j$, and then extract a prediction from the trained model for $I_j$. 

The final step is to determine the final aggregated answer for each case $R_{P_j}$, according to the prediction extracted for its associated instance $I_j$:
\begin{itemize}
    \item In AMP, each instance $I_j$ is given multiple binary-classification, one for each of the labels $\{O_{MR},O_{HAC},O_{WC},O_{SP},O_{DA}\}$, including a confidence measure (i.e., classification probability), for each predicted label value. The aggregation result for $R_{P_j}$ is given by applying the aggregation method, which its associated label prediction has the highest confidence measure. 
    \item In DAP, each instance $I_j$ is classified with the predicted to be the correct answer for the case. Thus, the returned answer is the one predicted by the classifier
\end{itemize}

\section{Results}

The main evaluation criterion we used was the success rate over the complete set of problems, which was calculated according to the outcomes of the aggregation procedures.
In this section, we describe the experimental results and compare the performances of our two aggregation approaches and the multiple possible aggregation methods.

\begin{figure}[ht]
\includegraphics[width=0.8\columnwidth]{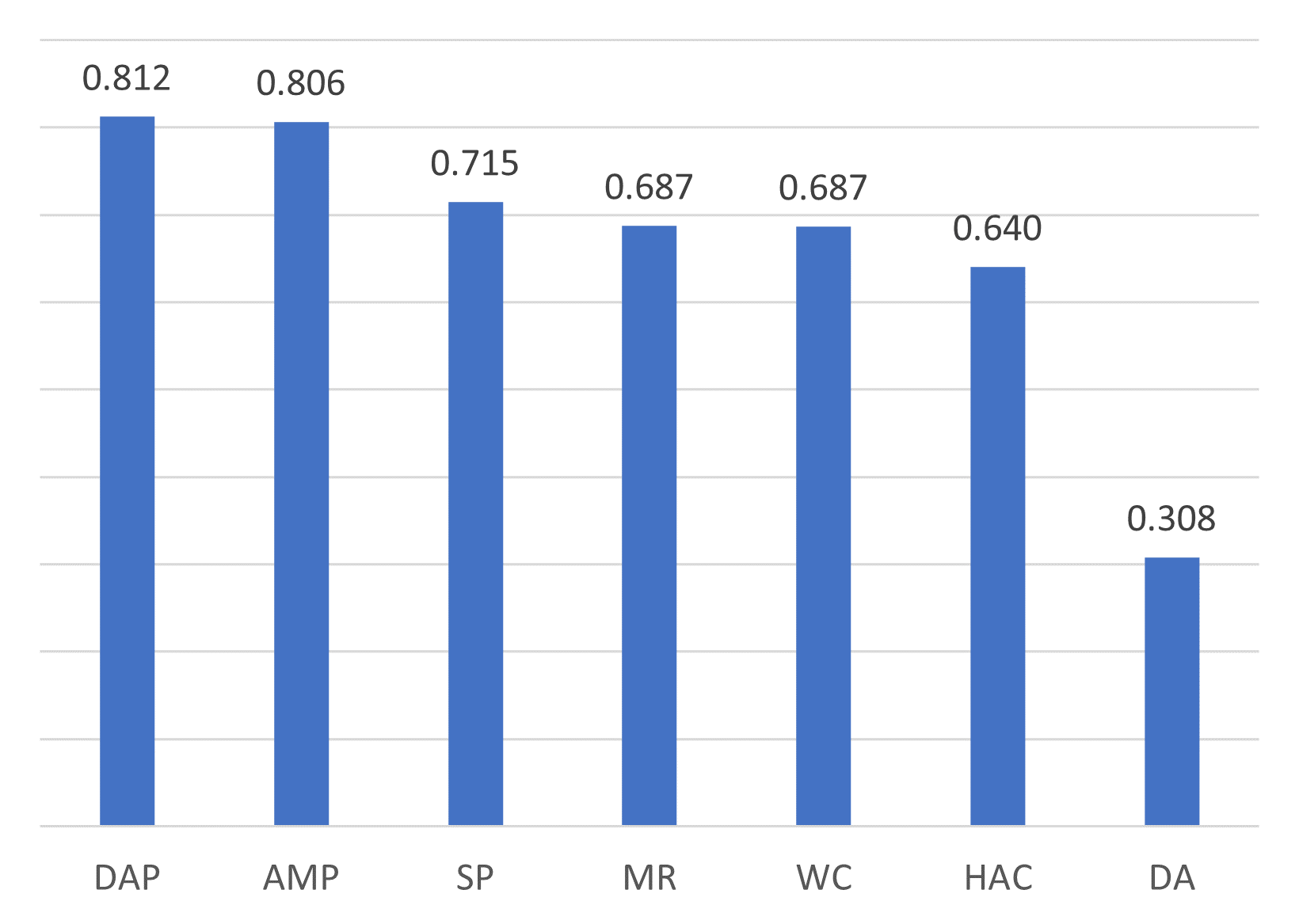}
\centering
\caption{Aggregation performance (the success rate) of the Direct-Answer-Prediction (DAP) approach, the Aggregation-Method-Prediction (AMP) approach, the Surprisingly Popular (SP) method,  the Majority Rule (MR), the Weighted Confidence (WC) method, the Highest Average Confidence (HAC) method, and the Devil's Advocate (DA) aggregator.}
\label{fig:main-res}
\end{figure}

Figure \ref{fig:main-res} presents the results of the evaluation, which provide the success rate (proportion of cases in which the correct answer was returned) of the AMP and DAP approaches, compared to the results of the SP, MR, WC, HAC and DA specific methods. 
As shown in Figure \ref{fig:main-res}, both of our proposed approaches increase the success rate significantly compared to each of the specific aggregation methods (for the results of the statistical tests, see Table \ref{tab:stat}, in the Appendix). 

\begin{figure}[ht]
\includegraphics[width=0.9\columnwidth]{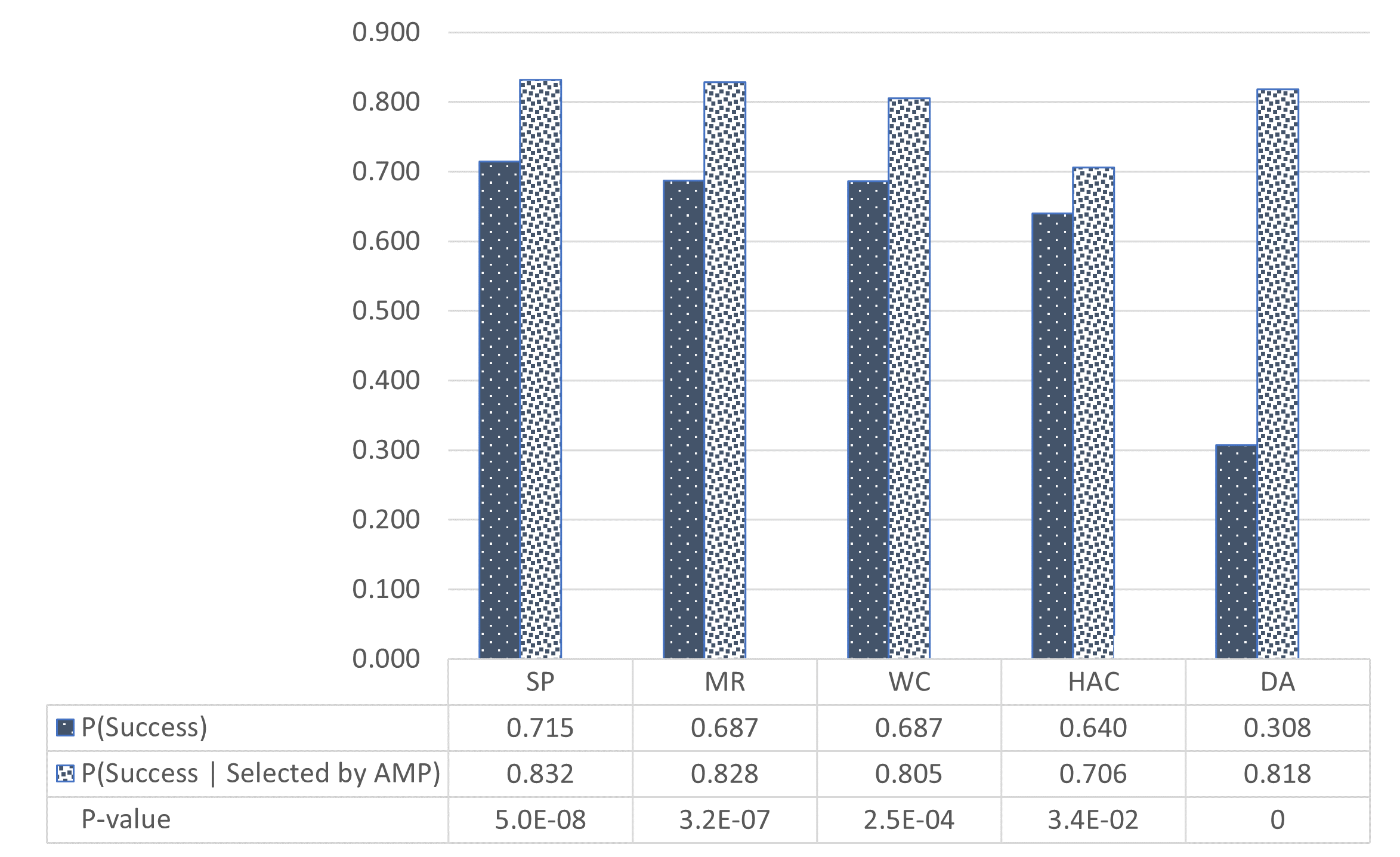}
\centering
\caption{Aggregation performance (proportion of cases where the correct answer was chosen) of Highest Average Confidence (HAC), Majority Rule (MR), the Devil's Advocate (DA), Surprisingly Popular (SP) and Weighted Confidence (WC) when 1) applied to the full data set (\textit{P(Success)}), and 2) chosen by the Aggregation-Method-Prediction (AMP) approach. (\textit{P(Success $|$ Selected by AMP)}).}
\label{fig:amp-select}
\end{figure}

\begin{figure}[ht]
\includegraphics[width=0.8\columnwidth]{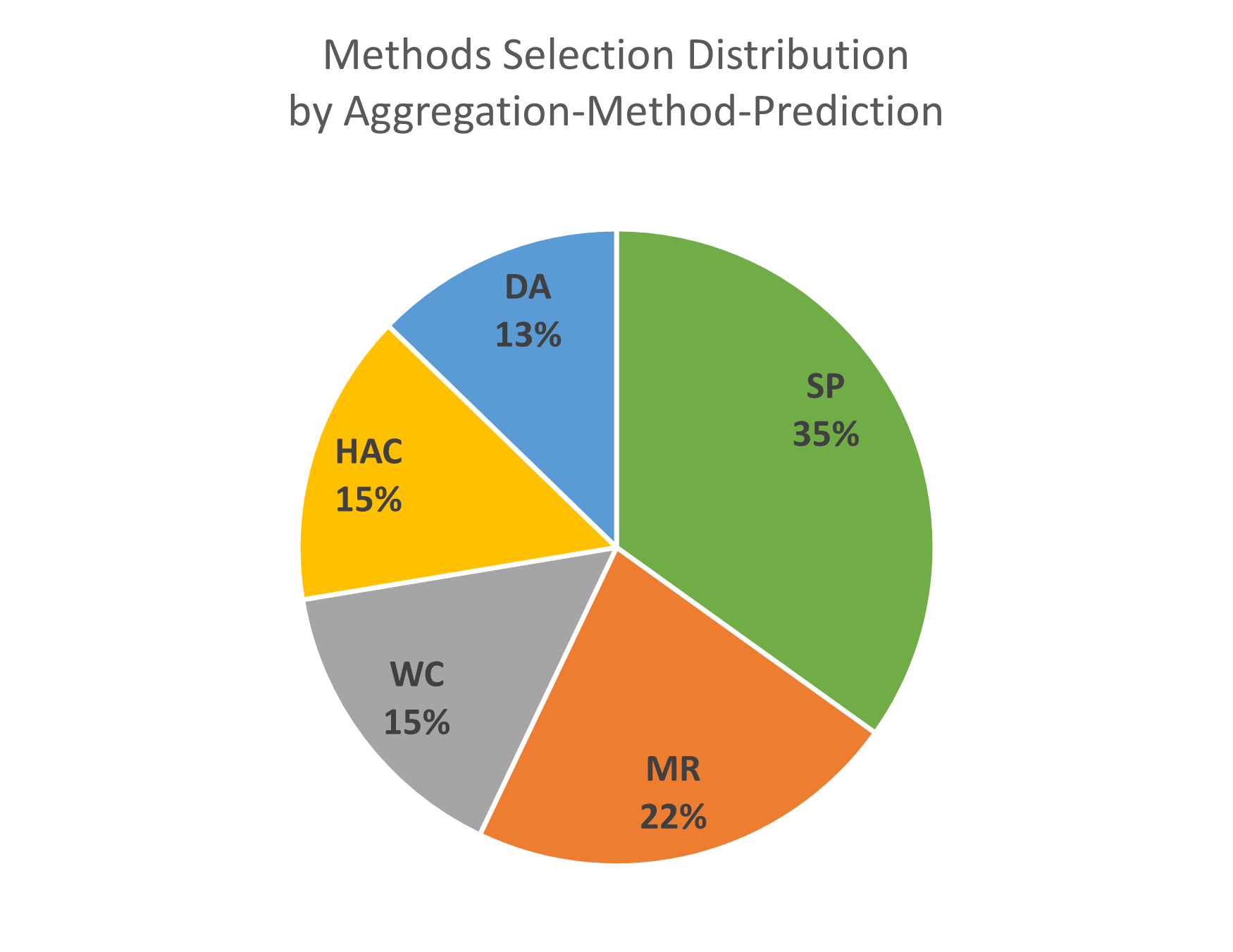}
\centering
\caption{The distribution of choices by the Aggregation-Method-Prediction (AMP) approach, showing the percentage of instances for which each aggregation method was chosen.}
\label{fig:amp-dist}
\end{figure}

One proof of a successful context-sensitive selection of the aggregation method as performed by the AMP approach would be reflected by an increase in the success rate of each method when applied to the cases in which that method was chosen by AMP, compared to its context-free success rate when applied indiscriminately (i.e., in uniform fashion) over the entire data set.
The distribution of AMP choices, i.e., the percentage of cases for which each of the specific aggregation methods was chosen, is depicted in Figure \ref{fig:amp-dist}. Figure \ref{fig:amp-select} presents the success rate of each aggregation method when applied to the cases in which the AMP approach chose that method (i.e., \textit{P(Success $|$ Selected by AMP)}), compared to the success rate of that method when applied to the entire data set (i.e., \textit{P(Success)}).
Comparing the success rate of each method in Figure \ref{fig:amp-select}, we can see a significant increase in the success rate when chosen by the AMP approach for each aggregation method, which can serve as solid evidence for the effectiveness of that approach.
Formally, for each aggregation method $AG_i$, the following applies: \\
 \textit{P(Success by $AG_i$)} $<$ \textit{P(Success by $AG_i | AG_i$ is selected by AMP)}.

\begin{figure}[ht]
\includegraphics[width=0.8\columnwidth]{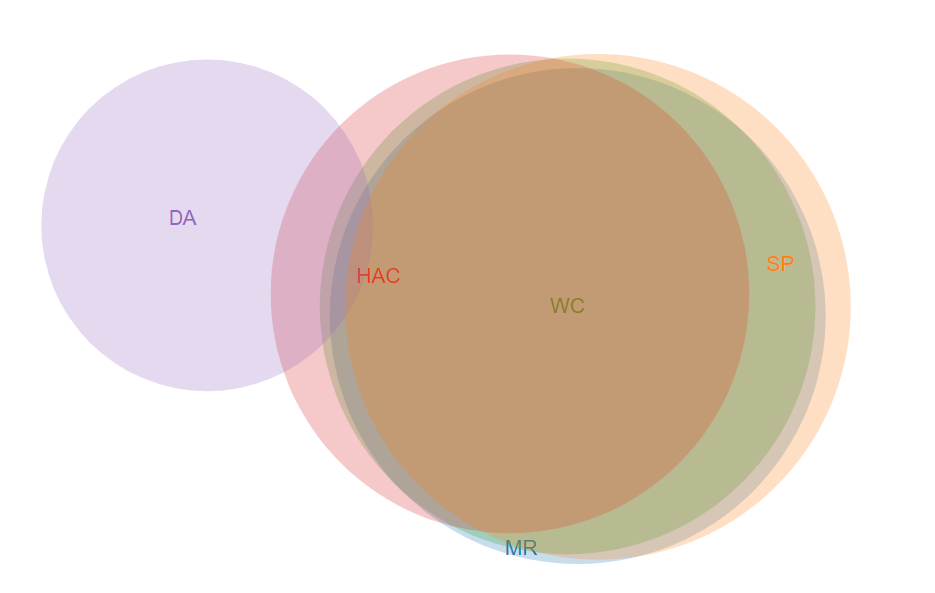}
\centering
\caption{Venn diagram depicting the cases resulted with a successful aggregation outcome when applying the methods: Surprisingly Popular (SP), Majority Rule (MR), Weighted Confidence (WC), Highest Average Confidence (HAC) and the Devil's-Advocate (DA) aggregator.}
\label{fig:venn}
\end{figure}

Figure \ref{fig:venn} depicts a Venn diagram that provides the interaction of the cases in which applying each aggregation method resulted in a successful aggregation outcome. Note that, with the inclusion of the DA aggregator, every case in the data set is solved (i.e., successfully aggregated) by at least one method. As observed from Figure \ref{fig:venn}, there are a considerable amount of cases (202 out 1209, to be exact) in which only the application of DA can result in making the correct decision. This observation allows a further understanding of the benefit of considering the DA aggregator, which is less apparent when observing only its success rate when applied indiscriminately (0.308, as depicted in Figure \ref{fig:main-res}). And in fact, in the 13\% of the cases in which the AMP approach selected the DA (as depicted in Figure \ref{fig:amp-dist}), its success rate was the highly respectable one of 0.818 (as illustrated in Figure \ref{fig:amp-select}).

\section{Ablation Experiments}

Our two approaches consist of learning features and labels that depend on the basic information given by each response, as described in Table \ref{tab:response-table}. For example, the value of the feature $f_{SP}$, used by the DAP approach, depends on the respondents providing their predicted support for each answer (i.e., $ps[a_i]$ for each $a_i \in A_P$). 

The implementation of our approach, as presented in the evaluation, assumes that each response contains the complete information, as shown in Table \ref{tab:response-table}. This assumption, however, might be a potential limitation since this information is not always available. Providing information beyond the basic information of the respondent's voted answer can be time-consuming and demanding for the worker and costly for the requester. Thus, it is crucial to determine - is it all worth it? Can we do without some of the excess information?

To answer these questions, we performed several ablation experiments for each aggregation approach. The ablation experiments were carried out by observing the results when eliminating different subsets of information from the input responses (e.g., confidence in the answer), and thus, also all of the engineered features and/or labels that depend on that information. 

Thus, we distinguish between three groups of features, from the features listed in Table \ref{tab:features}, according to the responses' information origin from which they are composed of:
\begin{itemize}
    \item \textit{Voting-features}: Features that are composed using (only) the vote component within the responses (i.e., $v$). This group composed of the following features' index: 1,2,3,4,5,18,19,20,21.
    \item \textit{Confidence-features}: Features that are composed using the reported confidence component within the responses (i.e., $c$). This group composed of the following features' index: 6,7,8,9,10,11,12,16,17,22,23,24,25,26,27. 
    \item \textit{Predicted-Support-features}: Features that are composed using the answers' support prediction component within the responses (i.e., $ps$ array). This group composed of the following features' index: 13,14,15,16,17.
\end{itemize}
 
Specifically, for the ablation experiments of the AMP approach, we tested the exclusion of the following feature sets and labels: \textit{Confidence-features}, \textit{Predicted-Support-features}, $\{O_{WC},O_{HAC}\}$, $O_{SP}$, $O_{DA}$. 
For the ablation experiments of the DAP approach, we tested the exclusion of the following: \textit{Confidence-features}, \textit{Predicted-Support-features}, $\{f_{WC},f_{HAC}\}$, $f_{SP}$, $f_{DA}$.

We consider the set of features included in the \textit{Voting-features}, as well as the label $O_{MR}$ for the AMP approach and $f_{MR}$ for the  DAP approach, to be the basic form of data since they are based on the information included in any collective decision-making case. Thus, these features and labels are included in every ablation experiment for both approaches, as the purpose is to examine the value of using the additional meta-cognitive information provided in the responses. 

Note that for each experiment in which one or more of the labels associated with the methods WC, HAC, and SP were removed (i.e., $O_{WC}$, $O_{HAC}$, $O_{SP}$ for the AMP approach; and $f_{WC}$, $f_{HAC}$, $f_{SP}$ for the DAP approach), the DA's labels (i.e., $O_{DA}$ for the AMP approach, and $f_{DA}$ for the DAP approach) were updated accordingly. Thus, if a method's label was excluded, the DA did not consider this method's choice in its aggregation procedure. 

\section{Ablation Results}

Table \ref{tab:abl-amp} and Table \ref{tab:abl-dap} presents the ablation experiments results for AMP and DAP approaches, respectively. The first row of both tables presents the approach's original results, obtained without excluding any component.

For the AMP approach, most of the ablation experiments' results did not differ significantly from the original results ($p>0.05$ using a two-tailed proportion test). 
There were only two cases with significantly different results, specifically a lower success rate than the original results. The first is the case in which all of the following components were excluded: \textit{Confidence-features}, \textit{Predicted-Support-features}, and the labels $O_{WC}$, $O_{HAC}$ and $O_{SP}$. Hence, there is a significant value in adding the meta-cognitive features.
The second case of a significant decrease in success rate occurred by removing the $O_{DA}$ label, i.e., by eliminating the DA aggregator from consideration by the AMP approach. Thus, having the option of choosing Devil's Advocate as an aggregator seems to be necessary.

As for the DAP approach, none of the results obtained in the ablation experiments deviated significantly from the original results ($p>0.05$ using a two-tailed proportion test). In other words, at least for problems that have only two possible answers, the \textit{Voting-features} subset seems to be sufficient for this method (see, however, the Discussion). Moreover, this approach appears to be more robust to changes compared to the AMP approach, as there is a lower performance variability when excluding different portions of the features.

\begin{table}[ht]
\caption{Aggregation-Method-Prediction: Ablation Results}
\begin{tabular}{ccccccc|c}
\textit{Voting-features} & \textit{Confidence-features} & \textit{Predicted-Support-features} & $O_{MR}$ & $O_{WC},O_{HAC}$ & $O_{SP}$ & $O_{DA}$   & Success rate   \\ \hline
V             & V          & V           & V        & V                   & V        & V        & 0.806          \\ \hline
V             &            & V           & V        & V                   & V        & V        & 0.800          \\ \hline
V             & V          &             & V        & V                   & V        & V        & 0.816          \\ \hline
V             &            &             & V        & V                   & V        & V        & 0.795          \\ \hline
V             & V          & V           & V        &                     & V        & V        & 0.808          \\ \hline
V             & V          & V           & V        & V                   &          & V        & 0.797          \\ \hline
V             & V          & V           & V        &                     &          & V        & 0.791          \\ \hline
V             & V          & V           & V        & V                   & V        &          & \textbf{0.754} \\ \hline
V             &            & V           & V        &                     & V        & V        & 0.806          \\ \hline
V             & V          &             & V        & V                   &          & V        & 0.799          \\ \hline
V             &            &             & V        &                     &          & V        & \textbf{0.763}         
\end{tabular}
\label{tab:abl-amp}
\end{table}

\begin{table}[ht]
\caption{Direct-Answer-Prediction: Ablation Results}
\begin{tabular}{ccccccc|c}
\textit{Voting-features} & \textit{Confidence-features} & \textit{Predicted-Support-features} & $f_{MR}$ & $f_{WC},f_{HAC}$ & $f_{SP}$ & $f_{DA}$ & Success rate   \\ \hline
V             & V          & V           & V        & V                 & V        & V        & 0.812          \\ \hline
V             &            & V           & V        & V                 & V        & V        & 0.812          \\ \hline
V             & V          &             & V        & V                 & V        & V        & 0.825          \\ \hline
V             &            &             & V        & V                 & V        & V        & 0.803          \\ \hline
V             & V          & V           & V        &                   & V        & V        & 0.804          \\ \hline
V             & V          & V           & V        & V                 &          & V        & 0.805          \\ \hline
V             & V          & V           & V        &                   &          & V        & 0.806          \\ \hline
V             & V          & V           & V        & V                 & V        &          & 0.813          \\ \hline
V             &            & V           & V        &                   & V        & V        & 0.811          \\ \hline
V             & V          &             & V        & V                 &          & V        & 0.815          \\ \hline
V             &            &             & V        &                   &          & V        & 0.798         
\end{tabular}
\label{tab:abl-dap}
\end{table}   

\section{Discussion}

As observed in the Results section, both our approaches, AMP and DAP, showed promising performances, significantly increasing success rates compared to all standard rule-based aggregation methods. Moreover, the AMP method demonstrated successful navigation ability between the various aggregation methods and, when choosing them, considerably increased their probability of producing optimal aggregation results.

The general concepts and principles underlying our approaches are not confined to specific domains and tasks and are adaptive to various situations (contexts) in an automated fashion. Thus, the implementation of our approaches can be generalized to any purpose whose end goal is to perform a successful aggregation process. Assuming sufficient training cases, the one-shot nature of these aggregation techniques allows a fast and easy application to each new collective decision-making case while being oblivious to the nature of the decision. That being said, the methodology can potentially be extended by including additional descriptive features, such as task type or group size. It can also be used alongside other techniques in the form of an ensemble of aggregators. 

There were different results and observations for the two approaches in the ablation experiments. For the DAP approach, we observed no significant changes in success rate compared to its original results. This can be interpreted as a somewhat positive outcome. When attempting to predict the correct answer, it sometimes may not be necessary to request additional information from the decision-makers beyond the voted answer. However, one should be cautious when applying such a conclusion to situations with \textit{more than two} possible answers since the basic information alone might be insufficient. 

Since an important limitation of the experimental evaluation was that it was performed over a set of binary-choice tasks, due to a large number of cases necessary, future work can test our approaches on an extensive collection of multiple-choice problems.  

We have, in fact, successfully applied two other machine-learning approaches, which also included meta-cognitive feature-engineering, to infer the optimal answer for problems with possibly more than two answer options \citep{shinitzky2022exploiting}. In the first approach, we trained a classifier that predicts the correctness of a single response. In the second approach, we trained a classifier that predicts the correctness of a single answer option. We showed the superiority of both approaches over standard aggregation methods \citep{shinitzky2022exploiting}.

Contrary to the DAP approach, the results of ablation experiments in the case of the AMP approach demonstrated that eliminating the use of the respondents' confidence and prediction regarding the answers' support (both features and classification targets) damaged the AMP's performance quality significantly. From that, we can deduce that the meta-cognitive information provides an added value, and it is essential to include at least some of its derived features and/or labels.
We also observed a significant decrease in the success rate of AMP when the label $O_{DA}$ was removed. This is proof of the importance of considering the Devil's Advocate (DA) aggregation method when the objective is to select the best aggregation method to apply.  

Another proof of the DA aggregator's significance was demonstrated by observing the interaction of the cases in which applying each aggregation method resulted in a successful aggregation outcome. By including the DA in the arsenal of aggregation methods, we observed that the percentage of cases that were successfully (correctly) aggregated by at least one method, increased from 83.3\% to 100\%. This means that if the application of the DA is an option, the potential for successful navigation between aggregation methods, as attempted by the AMP approach, is in theory unlimited, while not including it creates an upper bound of around 83\% for correctness using only standard aggregation methods. This observation emphasizes the importance of the DA, especially for the success of the AMP approach. 

Future work can further develop the Devil's Advocate aggregator, for example, by defining a confidence measure for the aggregation results of each input method. The confidence measure can be used to weigh the aggregation methods' decisions when determining the DA aggregator's chosen answer. This allows the DA aggregator to base its decision on an objective, consistent and quantifiable measure; it would provide a more elegant solution for tie-breaking cases and better suit decision-making problems with more than two answers options.

\section{Conclusion}

In this study, we presented two machine-learning-based aggregation approaches that exploit the context created by the various properties of the problem (task) and the associated responses. The first, AMP, uses a classification model to predict which aggregation method would be best to apply to a given case, and then applies the chosen method. The second approach, DAP, uses a classification model to directly predict the correct answer while also using the choices of each aggregation method as features. 
We offered a meta-cognitive feature-engineering approach for characterizing a collective decision-making case in a context-sensitive fashion. Contrary to other existing machine-learning-based aggregation methods, our methodology is independent of the crowd-specific composition, prior knowledge, and personal track records of the individual decision-makers or groups. Experimental results show a significant increase in the rate of successful aggregation using our proposed approaches, compared to the uniform (indiscriminate) application of each rule-based aggregation method.
Another contribution of our work is a novel method, the \textit{Devil's-Advocate} (DA) aggregator, which selects the option as far as possible from the choices made by other methods. The strength of the DA lies primarily in its ability to allow the consideration of alternatives that otherwise will be overlooked, which, in challenging decision-making tasks, might turn out to be the correct decisions. 
Evidence for that was observed in the AMP's ablation results, which showed a significant decrease in success rate when removing the DA from consideration when choosing which aggregation method to apply. 

\section*{Acknowledgments}

\bibliographystyle{unsrtnat}
\bibliography{references}  

\section{Appendix}

\begin{table}[h]
\caption{Aggregation-Method-Prediction: Model Selection Results}
\begin{tabular}{ll|llllllllll}
\multicolumn{1}{l}{Multi-Label} & Classifier & Fold 1         & Fold 2         & Fold 3         & Fold 4         & Fold 5         & Fold 6         & Fold 7         & Fold 8         & Fold 9         & Fold 10     \\ \hline
\multirow{4}{*}{BR}             & BNB        & 0.641          & 0.665          & 0.642          & 0.640          & 0.659          & 0.652          & 0.657          & 0.661          & 0.665          & 0.669        \\
                                & KNN        & 0.776          & 0.783          & 0.778          & 0.788          & 0.769          & 0.768          & 0.779          & 0.791          & 0.813          & 0.796       \\
                                & LR         & 0.759          & 0.772          & 0.767          & 0.767          & 0.748          & 0.750          & 0.767          & 0.779          & 0.788          & 0.781         \\
                                & RF         & \textbf{0.800} & \textbf{0.812} & \textbf{0.808} & \textbf{0.803} & \textbf{0.793} & 0.789          & \textbf{0.805} & \textbf{0.827} & 0.830          & \textbf{0.817} \\ \hline
\multirow{4}{*}{CC}             & BNB        & 0.631          & 0.645          & 0.630          & 0.621          & 0.639          & 0.624          & 0.633          & 0.642          & 0.637          & 0.645           \\
                                & KNN        & 0.774          & 0.789          & 0.774          & 0.784          & 0.768          & 0.763          & 0.778          & 0.788          & 0.810          & 0.789          \\
                                & LR         & 0.749          & 0.767          & 0.754          & 0.748          & 0.731          & 0.726          & 0.754          & 0.778          & 0.768          & 0.764       \\
                                & RF         & 0.790          & 0.809          & 0.796          & 0.792          & 0.777          & \textbf{0.790} & 0.801          & 0.808          & \textbf{0.836} & 0.809        \\ \hline
\multirow{4}{*}{LP}             & BNB        & 0.642          & 0.644          & 0.643          & 0.642          & 0.665          & 0.659          & 0.660          & 0.675          & 0.682          & 0.690       \\
                                & KNN        & 0.776          & 0.783          & 0.778          & 0.788          & 0.769          & 0.768          & 0.779          & 0.791          & 0.813          & 0.796          \\
                                & LR         & 0.767          & 0.784          & 0.772          & 0.777          & 0.756          & 0.757          & 0.774          & 0.798          & 0.799          & 0.788          \\
                                & RF         & 0.789          & 0.806          & 0.792          & 0.792          & 0.789          & 0.789          & 0.797          & 0.823          & 0.824          & 0.813           \\ \hline
\multicolumn{2}{c|}{Test Results}            & 0.851          & 0.793          & 0.826          & 0.835          & 0.934          & 0.917          & 0.860          & 0.645          & 0.479          & 0.592           
\end{tabular}
  \label{tab:cv-amp}
\end{table}

\begin{table}[h]
\caption{Direct-Answer-Prediction: Model Selection Results}
\begin{tabular}{l|llllllllll}
Classifier   & Fold 1         & Fold 2         & Fold 3         & Fold 4         & Fold 5         & Fold 6         & Fold 7         & Fold 8         & Fold 9         & Fold 10        \\ \hline
RF           & \textbf{0.793} & \textbf{0.804} & \textbf{0.802} & \textbf{0.792} & \textbf{0.792} & \textbf{0.781} & \textbf{0.802} & \textbf{0.824} & \textbf{0.827} & \textbf{0.820}   \\
LR           & 0.695          & 0.708          & 0.707          & 0.699          & 0.730          & 0.733          & 0.720          & 0.737          & 0.720          & 0.714           \\
KNN          & 0.763          & 0.767          & 0.757          & 0.760          & 0.757          & 0.759          & 0.765          & 0.790          & 0.797          & 0.782         \\ \hline
Test Results & 0.810          & 0.826          & 0.901          & 0.868          & 0.826          & 0.826          & 0.835          & 0.727          & 0.579          & 0.550         
\end{tabular}
  \label{tab:cv-dap}
\end{table}

\begin{table}[h]
\caption{Statistical Tests Results (P-Value)}
\begin{tabular}{l|ll|ll}
    & \multicolumn{2}{c|}{vs AMP} & \multicolumn{2}{c}{vs DAP} \\
    & McNemar Test         & Proportion Test         & McNemar Test      & Proportion Test     \\ \hline
SP  & 2.08E-14             & 6.13E-08                & 3.99E-13          & 8.18E-09            \\
MR  & 5.45E-19             & 8.18E-12                & 4.15E-18          & 6.72E-13            \\
WC  & 3.59E-20             & 6.08E-12                & 2.83E-18          & 4.92E-13            \\
HAC & 3.49E-26             & 0                       & 1.05E-25          & 0                   \\
DA  & 5.30E-84             & 0                       & 4.74E-90          & 0                  
\end{tabular}
\label{tab:stat}
\end{table}

\end{document}